\documentclass[runningheads]{llncs}

\usepackage{soul}
\usepackage{url}
\usepackage[hidelinks]{hyperref}
\usepackage[utf8]{inputenc}
\usepackage[small]{caption}
\usepackage{graphicx}
\usepackage{amsmath}
\usepackage{booktabs}
\usepackage{algorithm}
\usepackage{algorithmic}
\usepackage{multirow}
\usepackage{graphicx}
\usepackage{latexsym}
\usepackage{times}
\usepackage{xcolor}
\usepackage{colortbl}
\usepackage{graphicx}
\usepackage{enumitem}
\usepackage{algorithm}
\usepackage{algorithmic}
\usepackage{makecell}

\usepackage{amsthm}

\def\abbr{pFLFE}
\def\fname{Personalized Federated Learning via Feature Enhancement}

\begin{document}
\title{\abbr{}: Cross-silo Personalized Federated Learning via Feature Enhancement on Medical Image Segmentation}

%
%

\author{Luyuan Xie\inst{1,2}\and Manqing Lin\inst{1} \and Siyuan Liu\inst{1,2}\and ChenMing Xu\inst{1,2}\and Tianyu Luan \inst{3}\and Cong Li\inst{1,2}\and Yuejian Fang\inst{1,2}\thanks{Corresponding author: fangyj@ss.pku.edu.cn}\and Qingni Shen\inst{1,2}\thanks{Corresponding author: qingnishen@ss.pku.edu.cn}\and Zhonghai Wu\inst{1,2}}

%
\institute{$^{1}$School of Software and Microelectronics, Peking University, Beijing, China \\$^{2}$National Engineering Research Center for Software Engineering, Peking University, Beijing 100871, China \\$^{3}$State University of New York at Buffalo }
\maketitle              
\begin{abstract}


In medical image segmentation, personalized cross-silo federated learning (FL) is becoming popular for utilizing varied data across healthcare settings to overcome data scarcity and privacy concerns. However, existing methods often suffer from client drift, leading to inconsistent performance and delayed training. We propose a new framework, \fname{} (\abbr{}), designed to mitigate these challenges. \abbr{} consists of two main stages: feature enhancement and supervised learning. The first stage improves differentiation between foreground and background features, and the second uses these enhanced features for learning from segmentation masks. We also design an alternative training approach that requires fewer communication rounds without compromising segmentation quality, even with limited communication resources. Through experiments on three medical segmentation tasks, we demonstrate that \abbr{} outperforms the state-of-the-art methods. 
\end{abstract}
\section{Introduction}

Cross-silo Federated Learning (FL) \cite{xie2024mh,knnper,fedavg,tan2022federated} has recently achieved promising pro-gress in medical image segmentation tasks such as \cite{fedsm,li2019privacy,lcfed}. It trains the network to gather data information from all clients without actually accessing data . Considering strict privacy regulation requirements of medical data, exploring cross-silo FL in medical image segmentation is important and useful to real-world applications \cite{mtl2,li2021ditto,mtl1,mtl3}. To deal with data heterogeneous among clients caused by different medical protocols, personalized federated learning approaches such as \cite{fedper,fedhealth,yang2021federated,roth2021federated} have made significant progress by using an aggregation of centralized and client-specific network design \cite{sunet}.

However, despite the recent progress made in personalized FL \cite{karimireddy2020scaffold,fedprox,clustered1,clustered2}, previous works such as FedRep \cite{fedrep} and LG-FedAvg \cite{lg-fedavg} still suffer from client drifting problems in medical image segmentation tasks \cite{liu2021feddg,fedsm,lcfed}. The simple feature extraction and aggregation process would cause client drifting problems, which would degrade the performance and make the training process unstable. We show the training process of FedRep (\textcolor{red}{red line}) and LG-FedAvg (\textcolor{blue}{blue line}) in Fig.1 (a). It is easy to observe that both FedRep and LG-FedAvg are not stable in training, and their performance drops as the training converges. To analyze the reasons that cause this issue, we visualize the t-distributed stochastic neighbor embedding (t-SNE) figure of foreground and background samples of FedRep and LG-FedAvg features in Fig.1 (b) and Fig.1 (c). The figures show clear distribution overlappings of foreground samples (\textcolor{cyan}{cyan points}) and background samples (\textcolor{red}{red points}) for both FedRep and LG-FedAvg. The indistinguishable feature distributions of foreground and background samples make the following network harder to classify foreground and background pixels.

\begin{figure}[t]
    \centering
    \includegraphics[width=1\textwidth]{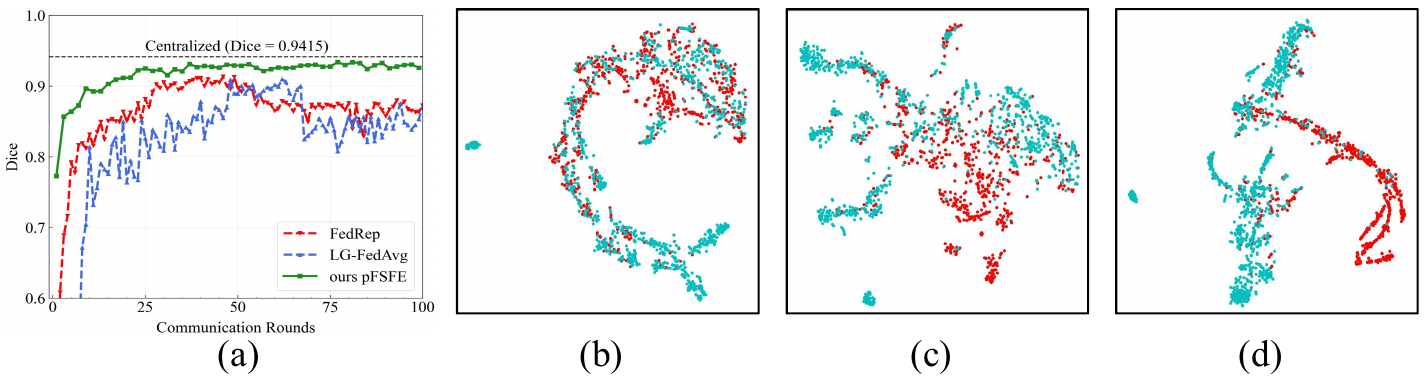}
    \caption{Previous personalized federated learning approaches suffer from client drifting problems. (a). Training progress of previous approaches (\textcolor{red}{red} and \textcolor{blue}{blue} line) compared with our approach (\textcolor{green}{green} line). (b)(c) Foreground (\textcolor{cyan}{cyan points}) and background (\textcolor{red}{red points}) feature distribution t-SNE for FedRep and LG-FedAvg. (d) Feature distribution t-SNE for our \abbr{}. The stability of previous works' training processes is not ideal and their foreground and background feature distribution are evidently overlapped. For comparison, \abbr{} has a more stable training process, evidently split foreground and background feature distribution, and better segmentation accuracy.}
    \label{fig:teaser}
\end{figure}

To address these challenges, we introduce a new framework, namely personalized cross-silo Federated medical image Segmentation via Feature Enhancement (\abbr{}), to enhance feature representation on each client without compromising data privacy. Our framework employs a self-supervised, contrastive approach for feature enhancement that relies solely on positive samples \cite{xie2023shisrcnet,wu2022federated,dong2021federated,wu2022distributed}, reducing the need for large batch sizes and avoiding the sharing of features between clients \cite{grill2020bootstrap,zhang2022contrastive,qi2022contrastive}. This approach ensures data privacy and is particularly designed for medical applications where data is often limited. 

Specifically, we design two federated learning frameworks including one with better performance and one that achieves comparable performance while using much fewer communication rounds. Our first framework consists of 4 main phases: local feature enhancement, local supervised training, and a global aggregation process after each local step, which together improve segmentation performance and training stability. The local supervised training will extract the information from each local client, and the local feature enhancement step will make personalized corrections to the shifted global parts after local training and aggregation. For our fast convergence framework, we use two similar local training steps but only one global aggregation process. The simplified communication design is found to have similar segmentation performance while converging much faster.

In summary, our contributions include the following:
\begin{itemize}[noitemsep,topsep=0pt]
\item We propose a novel personalized FL framework called  \abbr{} for medical image segmentation. \abbr{} tackles client drift problems in medical image segmentation FL with a feature enhancement network using only positive samples, which eliminates the requirements of negative samples or features from other clients.
\item We design an alternative fast-converging framework that can reach comparable performance in a few communication rounds, which is useful when communication resources are limited.

\item Our experiments on 3 segmentation tasks involving in total of 17 datasets show that, \abbr{} outperforms state-of-the-art results and achieves comparable performance with centralized learning, with high training stability and faster convergence.

\end{itemize}

\begin{figure}[t]
\centering
\includegraphics[width=1\textwidth]{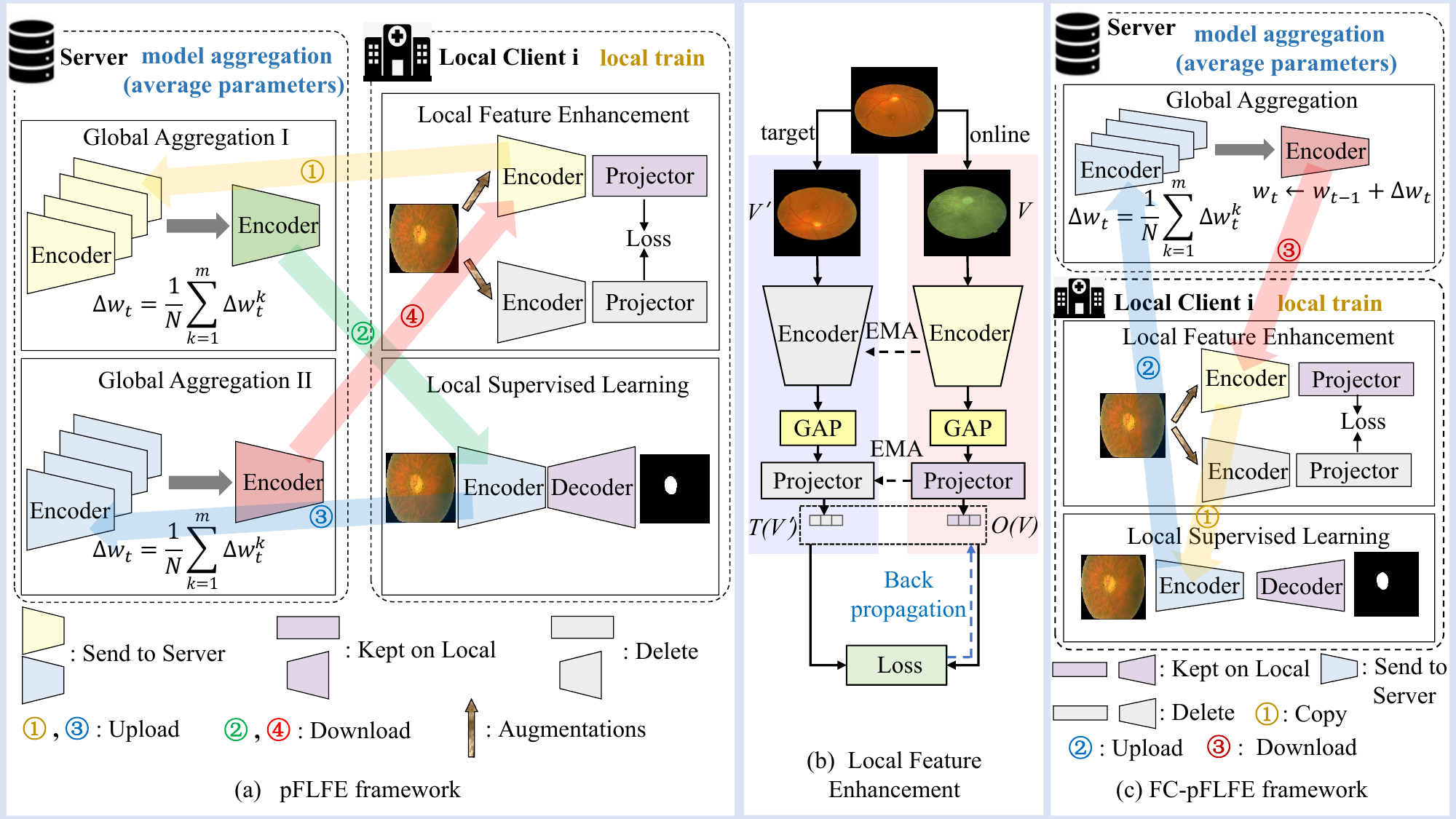}
\caption{The overview of (a) pFLFE framework, (b) Local Feature Enhancement and (c) FC-pFLFE framework.}
\end{figure}

\section{Methods}


\subsection{\abbr{} Framework}

In Fig.2 (a), we present the \abbr{} framework, consisting of 4 key stages: Local Feature Enhancement, Global Aggregation I, Local Supervised Learning, and Global Aggregation II, with stages 1 and 3 involving model uploads and stages 2 and 4 involving downloads. First, we enhance local feature extraction through a local feature enhancement module, which we will introduce details in Sec.~\ref{sec:fem}. Then, during Global Aggregation I, encoders from all clients are aggregated. 
Subsequently, Local Supervised Learning refines segmentation networks on individual clients, extracting information from local data with annotation.
Finally, Global Aggregation II shares refined encoders, further integrating supervised learning information across clients. In total, \abbr{} uniquely improves local feature extraction and personalization, enabling superior performance and faster convergence without requiring data sharing, thus maintaining data privacy. 

The components $E_i$, $D_i$, and $P_i$ represent the encoder, decoder, and projector of the $i$-th client in Fig.2 (a), respectively. $E_i$ is a shared parameter, and considering the issue of imbalanced client data and fairness in performance distribution \cite{li2021ditto,xu2021federated}, we adopt an average client weights approach rather than average data \cite{xu2021federated}. Due to the two-stage training of \abbr{}, optimization losses for client $i$ are as follows:
\begin{equation}
\begin{array}{cc}
     &{\cal L}_{1,i} = \mathcal{L}_{\textit{MSE}}\left(P_i(E_i(x^{a}_{i})), P^{\prime}_i(E^{\prime}_i(x^{a^{\prime}}_{i}))\right) + \mathcal{L}_{\textit{MSE}}\left(P_i(E_i(x^{a^{\prime}}_{i})), P^{\prime}_i(E^{\prime}_i(x^a_{i}))\right)\\
    &{\cal L}_{2,i} = \mathcal{L}_{\textit{Dice}}(D_i(E_i(x_{i})), y_{i})+\mathcal{L}_{\textit{CE}}(D_i(E_i(x_{i})), y_{i})
\end{array}
\label{eq:Lxx}
\end{equation}

${\cal L}_{1,i}$ and ${\cal L}_{2,i}$ are the loss functions of the first and second stages, respectively.
${\cal L}_{\textit{MSE}}$ means mean squared error loss, ${\cal L}_{\textit{CE}}$ is cross-entropy loss and ${\cal L}_{\textit{Dice}}$ is Dice loss \cite{fedsm}. $P^{\prime}_i(\cdot)$ and $E^{\prime}_i(\cdot)$ are another branch of client $i$ in first stage.
$x_{i}$ and $y_{i}$ are the data and labels on client $i$. $a$ and $a^{\prime}$ are two different types of data augmentation.

%





\subsection{Local Feature Enhancement}
\label{sec:fem}
As shown in Fig.2 (b), the Local Feature
Enhancement (LFE) consists of two branches: an online network $O$ and a target network $T$. The target network is initialized with the same parameters as the online network. 
We train both branches using only positive samples, eliminating the need for negative samples or features from other clients.
Given image $I$, we use two different data augmentation strategies for the same image. The resulting images are denoted as $V$ and $V'$, respectively. $V$ and $V'$ are passed through the encoder and projector in both the online and target branches, resulting in $O(V)$ and $T(V')$. In order for the encoder to learn the representation invariance of the same data, we 
calculate the mean squared error (MSE) between the normalized $O(V)$ and $T(V')$ like \cite{grill2020bootstrap，xie2024trls}:
\begin{equation}
    \mathcal{L} = \left\| \overline{O(V)}-\overline{T(V')} \right\|_2^2 =  2 - 2 \cdot \frac{{\left\langle O(V),T(V') \right\rangle }}{{{{\left\| O(V) \right\|}_2} \cdot {{\left\| {T(V')} \right\|}_2}}}
    \label{eq:Lxx}
\end{equation}
$\overline{O(V)}$ and $\overline{T(V')}$ are the $\ell_2$-normalize of $O(V)$ and $T(V')$, respectively. $\left\langle \cdot \right\rangle$ means dot product, and ${\left\| \cdot \right\|}_2$ represents the $\ell_2$-norm of the features. We separately feed {$V{'}$} to the online network and $V$ to the target network to compute $\mathcal{L}{'} = \| \overline{O(V')}-\overline{T(V)} \|_2^2$, respectively. The total loss function is defined as:
\begin{equation}
    \mathcal{L}_{total}  =  \mathcal{L}  + \mathcal{L}^{\prime}
    \label{eq:Lxx}
\end{equation}

The target network updates through Exponential Moving Average (EMA) \cite{grill2020bootstrap}. And we remove it after first training stage, which reduces the burden of local storage.  In online branch, we upload encoder and retain the projector after the first training stage. Local Feature Enhancement can make personalized corrections to the previous round second stage of training and aggregated encoder that generates drift in each round. 


\subsection{Fast Converging  \abbr{}} 

It is crucial to evaluate the communication overhead of federated learning. However, a complete training round of \abbr{} includes two communication processes. This significantly increases the communication cost. To mitigate this issue, we provide a fast converging version of \abbr{} named Fast Converging \abbr{} (FC-\abbr{}) as illustrated in Fig.2 (c),  which can provide comparable results with very few communication rounds.




In the FC-\abbr{} training phase, we begin with Local Feature Enhancement, which is the same feature enhancement procedure as in \abbr{}. Then, the trained encoder is passed to the local segmentation model for Local Supervised Learning. After that, the encoder is then uploaded to the server for aggregation. FC-\abbr{} only uses 1 aggregation step for 1 round of local feature enhancement and supervised learning, unlike \abbr{} which uses 2 aggregation steps.
This design allows us to reduce the number of communication rounds while largely preserving the feature enhancement and local knowledge acquisition of \abbr{}.

\begin{table}[t]
  \centering
  \caption{Test Dice coefficient comparison of optic disc/cup segmentation. “Client $k$ Local” refers to the model training locally only using data on client $k$. 
  We also report the Dice\textsubscript{ACli}, Dice\textsubscript{AImg} and VDice\textsubscript{ACli}. For Dice\textsubscript{ACli} and Dice\textsubscript{AImg}, large is better. For  VDice\textsubscript{ACli}, smaller is better. \textbf{Bold} numbers indicate the best except for the centralized method. \underline{Underlines} indicate the second bests. We can observe that our \abbr{} and FC-\abbr{} not only outperform SOTA in general, but also on most clients respectively. FT represents fine-tuning.}
  \resizebox{0.9\linewidth}{!}{
    \begin{tabular}{c|c|ccccccc|>{\columncolor{gray!20}}c>{\columncolor{gray!20}}c>{\columncolor{gray!20}}c}
    \hline
    \rowcolor{red!10} \multicolumn{2}{c|}{Model} & {{Client1}} & {{Client2}} & {{Client3}} & {{Client4}} & {{Client5}} & {{Client6}} & {{Client7}} & {{Dice\textsubscript{ACli}$\uparrow$}} & {{Dice\textsubscript{AImg}$\uparrow$}} & {{VDice\textsubscript{ACli}$\downarrow$}} \\
    \hline
    \multicolumn{2}{c|}{Centralized}  &0.9534 & 0.9425 & 0.9457 & 0.9532 & 0.9393 & 0.8987 & 0.9576 & 0.9415  & 0.9261  & 0.0185  \\
    \hline\hline
    \multicolumn{2}{c|}{Client1 Local}& 0.9172 & 0.4025 & 0.5314 & 0.0009 & 0.7076 & 0.3811 & 0.0446 & 0.4265  & 0.4698  & 0.3072  \\
    \multicolumn{2}{c|}{Client2 Local}& 0.7271 & 0.9001 & 0.8087 & 0.0002 & 0.7278 & 0.8218 & 0.5548 & 0.6486  & 0.7034  & 0.2830  \\
    \multicolumn{2}{c|}{Client3 Local}& 0.8483 & 0.8062 & 0.9054 & 0.0155 & 0.8524 & 0.8187 & 0.5174 & 0.6806  & 0.7428  & 0.2957  \\
    \multicolumn{2}{c|}{Client4 Local}& 0.0610 & 0.1299 & 0.0572 & 0.9322 & 0.0157 & 0.1299 & 0.4265 & 0.2503  & 0.1814  & 0.3055  \\
    \multicolumn{2}{c|}{Client5 Local}& 0.7904 & 0.6635 & 0.8329 & 0.0051 & 0.9153 & 0.7997 & 0.5529 & 0.6514  & 0.7358  & 0.2857  \\
    \multicolumn{2}{c|}{Client6 Local}& 0.8062 & 0.5368 & 0.3717 & 0.0037 & 0.5724 & 0.8815 & 0.6966 & 0.5527  & 0.6763  & 0.2746  \\
    \multicolumn{2}{c|}{Client7 Local}& 0.5465 & 0.8310 & 0.7339 & 0.0709 & 0.6579 & 0.6412 & 0.9242 & 0.6294  & 0.6113  & 0.2559  \\
    \hline\hline
    \multirow{6}{*}{Single model}&FedAvg & 0.9226 & 0.8863 & 0.9274 & 0.2009 & 0.9121 & 0.8669 & 0.8404 & 0.7938  & 0.8247  & 0.2438  \\
    &FedAvg+FT & 0.9379 & 0.9099  & 0.9246 & 0.5705 & 0.9037 & 0.8637 & 0.9171 & 0.8611  & 0.8617  & 0.1205  \\
    &SCAFFOLD & 0.8970 & 0.8703  & 0.9157 & 0.4001 & 0.8993 & 0.8975 & 0.8372  & 0.8167  & 0.8487  & 0.1717  \\
    &SCAFFOLD+FT & 0.9281 & 0.8833 & 0.9097  & 0.7855 & 0.8981 & 0.8796 & 0.8524 & 0.8767  & 0.8802  & 0.0433  \\
    &FedProx & 0.9123 & 0.8824 & 0.8567 & 0.3301 & 0.8726 & 0.8821 & 0.8521  & 0.7983  & 0.8306  & 0.1920  \\
    &FedProx+FT & 0.9301 & 0.8620 & 0.9273 & 0.7001 & 0.8801 & 0.8509 & 0.8743 & 0.8607  & 0.8567  & 0.0714  \\
    \hline
    \multirow{6}{*}{Presonalized FL} & Ditto & 0.6747 & 0.8830 & 0.9074 & 0.8335 & 0.8460 & 0.7350 & 0.9332 & 0.8304  & 0.7880  & 0.0869  \\
    &APFL  & 0.8762 & 0.8204  & \underline{0.9338} & 0.8261 & 0.7253 & 0.7732 & 0.8341 & 0.8270  & 0.7873  & 0.0623  \\
    &LG-FedAvg    & 0.9530 & 0.8420 & 0.8917 & 0.9237 & 0.9234 & 0.6138 & 0.9486 & 0.8709  & 0.7909  & 0.1106  \\
    &FedRep & 0.9342 & \underline{0.9408} & \textbf{0.9449} & 0.7134 & 0.9161 & 0.8702 & 0.7873 & 0.8724  & 0.8751  & 0.0830  \\
    &FedSM & 0.8995 & 0.8972 & 0.9216 & 0.8602 & 0.8251 & 0.8923 & 0.8659 & 0.8803  & 0.8733  & 0.0297  \\
    &LC-Fed & 0.9173 & \textbf{0.9422} & 0.9149 & 0.8041 & 0.9022 & 0.8761 & 0.7992 & 0.8794  & 0.8805  & 0.0525  \\
    \hline\hline
    \multirow{2}{*}{Ours} & { \abbr{}} & \textbf{0.9661} & 0.9185 & 0.9317 & \textbf{0.9381} & \underline{0.9300}  & \textbf{0.9002} & \textbf{0.9661} & \textbf{0.9358}  & \textbf{0.9232}  & \textbf{0.0222}  \\
    &{FC-\abbr{}} & \underline{0.9633} & 0.8919 & 0.9072 & \underline{0.9354} & \textbf{0.9376} & \underline{0.8933} & \underline{0.9630} & \underline{0.9274}  & \underline{0.9194}  & \underline{0.0282}  \\
    \bottomrule
    \hline
    \end{tabular}%
  \label{tab:addlabel}%
  }
\end{table}%

\section{Experiment Setup}

\textbf{Task and Dataset.} We evaluate our proposed  \abbr{} on 3 tasks. 
\noindent\textbf{Optic Disc/Cup segmentation:} We use 7 datasets from \cite{fedsm,lcfed}. Each dataset represents a client, so there are 7 clients. 
\noindent\textbf{Polyp segmentation:} The endoscopic images dataset is collected and annotated from four different centers \cite{Zhou2014,Tajbakhsh2015,Bernal2015,Jha2020}, and each center's dataset is treated as a separate client. This task has 4 clients.
\noindent\textbf{Prostate segmentation:} We use MRI data collected and annotated by 6 institutions \cite{liu2021feddg}. Each institution's dataset is treated as a client, resulting in 6 clients for this task.
Considering that some clients have limited data, we employed a widely used 1:1 train-test split for optic disc/cup and polyp segmentation. The train-test split protocol follows the standard split used in the latest work for prostate segmentation.

\noindent\textbf{Implementation Details.} To ensure reliability of our experiments, we employ a five-fold cross-validation and report the mean Dice coefficient \cite{fedsm}. We calculate the average Dice coefficient for each client ({Dice\textsubscript{ACli}), the average Dice coefficient for all test images ({Dice\textsubscript{AImg}), and the variance of Dice across clients (VDice\textsubscript{ACli}) to evaluate the model's performance and client discrepancy. (More details in supplementary materials.)


\begin{table}[t]
\begin{minipage}[h]{0.63\textwidth}
\centering
\captionof{table}{Test Dice coefficient comparison of polyp segmentation. FT represents fine-tuning.}
  \resizebox{1\linewidth}{!}{
    \begin{tabular}{c|c|cccc|>{\columncolor{gray!20}}c>{\columncolor{gray!20}}c>{\columncolor{gray!20}}c}
    \hline
   \rowcolor{green!10} \multicolumn{2}{c|}{Model} & {{Client1}} &{{Client2}} & {{Client3}} & {{Client4}} & {{Dice\textsubscript{ACli}$\uparrow$}} & {{Dice\textsubscript{AImg}$\uparrow$}} & {{VDice\textsubscript{ACli}$\downarrow$}}  \\
     \hline
    \multicolumn{2}{c|}{Centralized}  & 0.8051 & 0.6910 & 0.8620 & 0.7765 & 0.7837  & 0.7977  & 0.0617  \\
     \hline\hline
    \multicolumn{2}{c|}{Client1 Local} & 0.7633 & 0.1775 & 0.3398 & 0.3119 & 0.3981  & 0.3861  & 0.2196  \\
    \multicolumn{2}{c|}{Client2 Local}& 0.2177 & 0.5332 & 0.2550 & 0.3744 & 0.3451  & 0.3280  & 0.1231  \\
    \multicolumn{2}{c|}{Client3 Local}& 0.2513 & 0.1212 & 0.8234 & 0.4208 & 0.4042  & 0.4771  & 0.2643  \\
    \multicolumn{2}{c|}{Client4 Local}& 0.3477 & 0.3012 & 0.5053 & 0.7357 & 0.4725  & 0.5649  & 0.1698  \\
     \hline\hline
    \multirow{6}{*}{Single model}&FedAvg & 0.5249 & 0.4205 & 0.5676 & 0.5500  & 0.5158  & 0.5390  & 0.0570  \\
    &FedAvg+FT & 0.6047 & 0.4762 & 0.7513 & 0.6681 & 0.6251  & 0.6632  & 0.1005  \\
   & SCAFFOLD & 0.5244 & 0.3591 & 0.5935 & 0.5713 & 0.5121  & 0.5504  & 0.0918  \\
    &SCAFFOLD+FT & 0.5937 & 0.4312 & 0.8231 & 0.7208 & 0.6422  & 0.7014  & 0.1464  \\
    &FedProx & 0.5529 & 0.4674 & 0.5403 & 0.6301 & 0.5477  & 0.5770  & 0.0577  \\
    &FedProx+FT & 0.7441 & 0.5701 & 0.7438 & 0.6402 & 0.6746  & 0.6809  & 0.0737  \\
      \hline
    \multirow{6}{*}{Presonalized FL} &Ditto & 0.5720 & 0.4644 & 0.6648 & 0.6416 & 0.5857  & 0.6201  & 0.0779  \\
    &APFL  & 0.6120 & 0.5095 & 0.6333 & 0.5892 & 0.5860  & 0.5984  & \textbf{0.0468}  \\
    &LG-FedAvg  & 0.6053 & 0.5062 & 0.7371 & 0.5596 & 0.6021  & 0.6124  & 0.0855  \\
    &FedRep & 0.5809 & 0.3106 & 0.7088 & 0.7023 & 0.5757  & 0.6479  & 0.1613  \\
   & FedSM & 0.6894 & 0.6278 & 0.8021 & 0.7391 & 0.7146  & 0.7381  & 0.0641  \\
    &LC-Fed & 0.6233 & 0.4982 & 0.8217 & \underline{0.7654} & 0.6772  & 0.7325  & 0.1261  \\
 \hline\hline
    \multirow{2}{*}{Ours} &{ \abbr{}} & \underline{0.7895} & \textbf{0.7447} & \textbf{0.8799} & \textbf{0.7705} & \textbf{0.7962}  & \textbf{0.8021}  & \underline{0.0509}  \\
    &{FC-\abbr{}} & \textbf{0.7962} & \underline{0.7182} & \underline{0.8729} & 0.7653 & \underline{0.7882}  & \underline{0.7965}  & 0.0563  \\
    \bottomrule
    \hline
    \end{tabular}%
  \label{tab:addlabel}%
  }
    \vspace{-3pt}
\caption{Test Dice coefficient comparison of prostate segmentation. FT represents fine-tuning.}
  \resizebox{1\linewidth}{!}{
    \begin{tabular}{c|c|cccccc>{\columncolor{gray!20}}c>{\columncolor{gray!20}}c>{\columncolor{gray!20}}c}
    \toprule
    \hline
    \rowcolor{blue!10} \multicolumn{2}{c|}{Model} & Client1 & Client2   & Client3   & Client4 & Client5 & Client6   & {{Dice\textsubscript{ACli}$\uparrow$}} & {{Dice\textsubscript{AImg}$\uparrow$}} & {{VDice\textsubscript{ACli}$\downarrow$}} \\
    \midrule
    \hline
    
    \multicolumn{2}{c|}{Centralized Training} & 0.8157 & 0.8369 & 0.8678 & 0.8324 & 0.8469 & 0.7601 & 0.8266  & 0.8266  & 0.0336  \\
     \hline\hline
    \multicolumn{2}{c|}{Client1 Local} & 0.8062 & 0.0707 & 0.7680 & 0.0048 & 0.0520 & 0.0208 & 0.2871  & 0.2115  & 0.3544  \\
    \multicolumn{2}{c|}{Client2 Local}   & 0.2585 & 0.7802 & 0.4863 & 0.0928 & 0.4498 & 0.4810 & 0.4248  & 0.4033  & 0.2128  \\
    \multicolumn{2}{c|}{Client3 Local}    & 0.3342 & 0.0890 & 0.8045 & 0.0243 & 0.3769 & 0.1228 & 0.2920  & 0.2324  & 0.2624  \\
    \multicolumn{2}{c|}{Client4 Local} & 0.1760 & 0.1933 & 0.3115 & 0.6673 & 0.3448 & 0.2514 & 0.3241  & 0.3589  & 0.1647  \\
    \multicolumn{2}{c|}{Client5 Local} & 0.1062 & 0.5092 & 0.3118 & 0.4808 & 0.8488 & 0.4483 & 0.4509  & 0.4898  & 0.2239  \\
    \multicolumn{2}{c|}{Client6 Local}   & 0.2853 & 0.3701 & 0.1764 & 0.0693 & 0.4684 & 0.5361 & 0.3176  & 0.2973  & 0.1612  \\
     \hline\hline
     \multirow{6}{*}{Single model}
     &FedAvg & 0.5437 & 0.5881 & 0.7942 & 0.1412 & 0.7715 & 0.6364 & 0.5792  & 0.5244  & 0.2159  \\
     &FedAvg+FT & 0.7373 & 0.7174 & 0.8274 & 0.5703 & 0.8569 & 0.6784 & 0.7313  & 0.7142  & 0.0948  \\
     &SCAFFOLD & 0.5902 & 0.5312 & 0.7969 & 0.2888 & 0.7595 & 0.6349 & 0.6003  & 0.5548  & 0.1669  \\
     &SCAFFOLD+FT & 0.6841 & 0.6607 & 0.8122 & 0.6423 & 0.8012 & 0.6533 & 0.7090  & 0.6980  & 0.0703  \\
     &FedProx & 0.5261 & 0.4467 & 0.7521 & 0.1743 & 0.7640 & 0.5219 & 0.5309  & 0.4854  & 0.1990  \\
     &FedProx+FT & 0.7745 & 0.7392 & 0.8321 & 0.5122 & 0.7823 & \underline{0.7091} & 0.7249  & 0.6967  & 0.1024  \\
    \midrule
    \hline
    \multirow{6}{*}{Presonalized FL} &Ditto & 0.7677 & 0.7831 & \underline{0.8385} & 0.6933 & 0.8395 & 0.3693 & 0.7152  & 0.7338  & 0.1624  \\
    &APFL  &  0.7233 &  0.8002 &  0.7143 &  0.5342 & 0.7976   &  0.5314 & 0.6835 
 & 0.6850  & 0.1115 \\
    &LG-FedAvg    & 0.7357 & 0.7252 & 0.7558 & 0.7130 & 0.8041 & 0.3431 & 0.6795  & 0.7053  & 0.1532  \\
    &Fedrep & 0.7320 & 0.7517 & 0.7915 & 0.7050 & 0.7659 & \textbf{0.7379} & 0.7473  & 0.7384  & \textbf{0.0272}  \\
    &FedSM &    0.7643  &  0.8113 & 0.7503
 &  0.7155 &  0.7859& 0.6029 & 0.7384 & 0.7469 
  & 0.0674  \\
    &LC-Fed & 0.7124 &0.8259 &0.8231 & 0.7752 & 0.8011 & 0.5124
& 0.7417  & 0.7592  & 0.1094 
 \\
     \hline\hline
    \multirow{2}{*}{Ours} & \abbr{} & \underline{0.8300}  & \textbf{0.8604} & \textbf{0.8488} & \underline{0.7635} & \textbf{0.8766} & 0.6940 & \textbf{0.8122}  & \textbf{0.8145}  & \underline{0.0639}  \\
    &FC-\abbr{} & \textbf{0.8449} & \underline{0.8279} & 0.8177 & \textbf{0.8068} & \underline{0.8617} & 0.5848 & \underline{0.7906}  & \underline{0.8055}  & 0.0938  \\
    \bottomrule
    \hline
    \end{tabular}%
  \label{tab:addlabel}%
  }

\end{minipage}
\begin{minipage}[c]{0.36\textwidth}
\centering
 \captionof{table}{Abation studies and the number of personalized layers. Only last layer: only use the last layer as the personalized part.}
    \resizebox{1\linewidth}{!}{
    \begin{tabular}{c|cccc|>{\columncolor{gray!20}}c}
    \toprule
      Method   & Client1 & Client2 & Client3 & Client4 & {{Dice\textsubscript{ACli}$\uparrow$}}  \\
    \midrule
    \textbf{ \abbr{}} & \textbf{0.7895}  & \textbf{0.7447}  & \textbf{0.8799}  & \textbf{0.7705}  & \textbf{0.7962}  \\
    w/o LFE & 0.7234 & 0.4373 & 0.8346 & 0.6954 & 0.6727  \\
    only last layer & 0.7053 & 0.6154 & 0.7902 & 0.7043 & 0.7038  \\
    \midrule
    \multicolumn{4}{c|}{\makecell{the number of \\ personalized layers}} & Para(M)$\downarrow$ & {{Dice\textsubscript{ACli}$\uparrow$}} \\
    \midrule
    \multicolumn{4}{c|}{only last layer}         & 31.05 & 0.7038 \\
    \multicolumn{4}{c|}{last 3 layer}         & 31.01 & 0.6744 \\
    \multicolumn{4}{c|}{last 9 layer}         & 30.76 & 0.7893 \\
    \multicolumn{4}{c|}{last 14 layer}        & 29.74 & 0.7341 \\
    \multicolumn{4}{c|}{last 19 layer}        & 25.67 & 0.7654 \\
    \multicolumn{4}{c|}{All decoder} & \textbf{18.85} & \textbf{0.7962}  \\
    \bottomrule
    \end{tabular}%
  \label{tab:addlabel}%
  }
  \centering
  \caption{The KL divergence of personalized Federated learning framework with parameter decouple in three tasks. (LFE: Local Feature Enhancement)}
  \resizebox{\linewidth}{!}{
    \begin{tabular}{c|ccc|>{\columncolor{gray!20}}c}
    \toprule
          Task & optic disc/cup & polyp & prostate & {{Average KL$\downarrow$}}\\
   \hline\hline
    FedRep & 0.5012  & 0.7310  & 0.7710  & 0.6677  \\
    LG-FedAvg & 0.4923  & 0.6512  & 0.8512  & 0.6649  \\
    LC-Fed & 0.4631  & 0.6012  & 0.7412  & 0.6019  \\
    \hline\hline
     \makecell{\abbr{} \\ (w/o LFE)} & 0.5542  & 0.7632  & 0.8932  & 0.7369  \\

    \abbr{} & \textbf{0.4242}  & \textbf{0.5227}  & \textbf{0.5427}  & \textbf{0.4965}  \\
    \bottomrule
    \end{tabular}%
  \label{tab:addlabel}%
  }
    \vspace{4pt}
    \centering
  \caption{The impact of different models on  \abbr{} in the polyp task}
  \resizebox{\linewidth}{!}{
    \begin{tabular}{c|rrrr|>{\columncolor{gray!20}}c}
    \toprule
     Model &{Client1} & {Client2} & {Client3} &{Client4} & {{Dice\textsubscript{ACli}$\uparrow$}}  \\
    \midrule
    FCN   & 0.7913  & 0.5056  & 0.6058  & 0.7717 & 0.7086 \\
    Unet  & 0.7895  & 0.7447  & 0.8799  & 0.7705  & 0.7962 \\
    Unet++ & \textbf{0.8237}  & \textbf{0.7563}  & 0.8854  & \textbf{0.7926}  & \textbf{0.8145} \\
    Res-Unet & 0.8103  & 0.7252  & \textbf{0.8891}  & 0.7831  & 0.8019 \\
    \bottomrule
    \end{tabular}%
  \label{tab:addlabel}%
  } 
\end{minipage}

\end{table}

\begin{figure*}[t]
\centering
\includegraphics[width=0.9\textwidth]{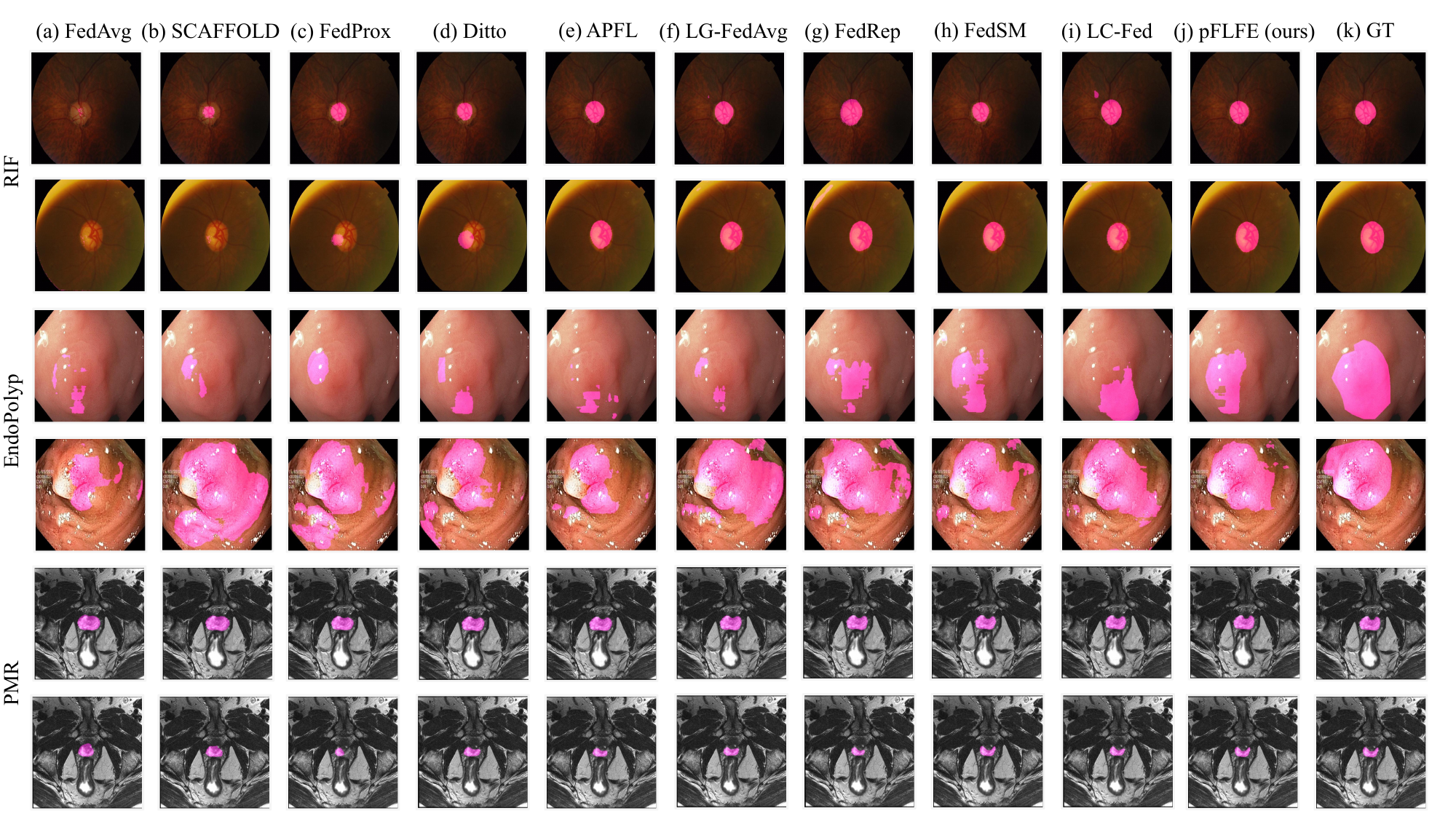}
\caption{Visualized comparison of personalized methods on three datasets. From each dataset, we randomly select two samples from different clients to form the visualization. (a-j)  Segmentation results by models trained with FedAVG,  SCAFFOLD, FedProx, Ditto, APFL, LG-FedAvg, FedRep, FedSM, LC-Fed and our method  \abbr{}; (k) Ground truths (denoted as ‘GT’);}
\label{fig6} 
\end{figure*}

\section{Results and Discussion}

\noindent\textbf{State-of-the-art comparison.} Table 1, Table 2, and Table 3 present the comparisons between  \abbr{} and previous approaches on 3 tasks. We can easily observe that our  \abbr{} achieves optimal performance compared to other FL frameworks in all three tasks. And the performance of  \abbr{} is close to that of centralized learning, even surpassing centralized learning in polyp segmentation task. The FC-\abbr{} performs slightly lower than  \abbr{} in all three tasks.

\noindent\textbf{Ablation experiment.} In Table 4, we conduct ablation experiments on  \abbr{} and adjusted the number of personalized layers of  \abbr{}. The results show that both LFE and the number of personalized layers have a significant impact on the performance of  \abbr{}. We use all decoders as personalized layers. This not only has performance advantages but also reduces the amount of parameters transmitted from the client to the server.

\noindent\textbf{Feature distribution.} In order to evaluate the quality of our enhanced feature, we compare the KL divergence of the feature distribution of the foreground and background samples. Smaller KL divergence indicates the global feature of foreground and background are better separated. As shown in Table.5,  \abbr{} feature has a smaller KL divergence than our baseline and previous works on all 3 tasks. These results verify the effectiveness of our feature enhancement design and the quality of our global feature.


\noindent\textbf{The impact of different models.} To demonstrate the impact of different models on  \abbr{}, we use different encoder-decoder segmentation networks on  \abbr{}. In Table 6, results show that using more complex models such as Unet++ and Res-Unet can help improve the performance of  \abbr{}.

\noindent\textbf{Visualized comparison.} Fig.3 visually compares the segmentation results on three tasks produced by our framework and baseline. Compared to other FL frameworks,  \abbr{} has better segmentation performance and is closer to ground truths.

\begin{figure*}[t]
\centering
\includegraphics[width=1\textwidth]{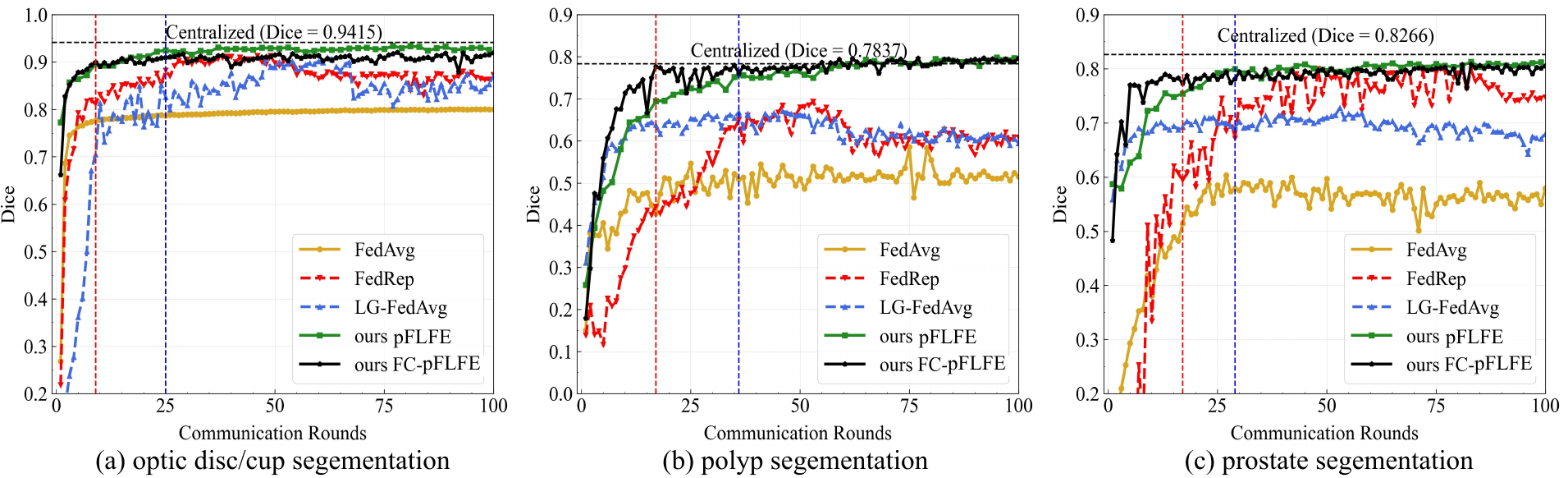}
\caption{Training progress of our  \abbr{} compared with previous results on 3 tasks. The \textcolor{green}{green} and black line is our  \abbr{} and FC-\abbr{} training progress, the \textcolor{red}{red}, \textcolor{blue}{blue}, and \textcolor{yellow}{yellow} lines are training progress of previous approaches. The black dashed line is the result of the centralized method. It is easy to observe that our  \abbr{} has better performance and more stable, faster-converged training progress. The FC-\abbr{} (\textcolor{red}{red} vertical line) reaches the near-optimal solution is significantly lower than that of  \abbr{} (\textcolor{blue}{blue} vertical line).}
\label{fig:traditional}
\end{figure*}

\noindent\textbf{The performance stability and convergence.} Better performance stability and convergence are beneficial for subsequent model selection and reducing communication rounds. We compare FedAvg and other pFL frameworks (FedRep and LG-FedAvg) in Fig.4. The experimental results show that \abbr{} has better convergence.  During the communication process,  \abbr{} exhibits less performance fluctuations. This proves its good performance stability. Meanwhile, We compare FC-\abbr{} with  \abbr{}. The results show that the number of communication rounds required for FC-\abbr{} (red line) to reach the near-optimal solution is significantly lower than that of  \abbr{} (blue line).

\begin{table}[ht]
  \centering
  \caption{Comparison of federated domain generalization results on Optic Disc/Cup segmentation.}
   \resizebox{0.65\linewidth}{!}{
    \begin{tabular}{c|ccccccc|>{\columncolor{gray!20}}c}
    \toprule
      \rowcolor{red!10} {unseen}&{{Client1}} & {{Client2}} & {{Client3}} & {{Client4}} & {{Client5}} & {{Client6}} & {{Client7}} & {{Dice\textsubscript{ACli}$\uparrow$}}\\
    \hline\hline
    FedAvg & 0.9281 & 0.8999 & 0.9098 & 0.7888 & 0.8823 & 0.8257 & 0.8968 & 0.8759 \\
    SCAFFOLD  & 0.9127 & 0.8965 & 0.9122 & \textbf{0.8855} & 0.8394 & 0.8496 & 0.8923 & 0.8840  \\
    FedProx  & 0.9031 & 0.8445 & 0.9273 & 0.8701 & 0.8905 & 0.8042 & 0.8905 & 0.8757\\
    Ditto  & 0.9099 & \textbf{0.9173} & 0.9221 & 0.8744 & 0.8523 & 0.8003 & 0.9056 & 0.8831  \\
    FedRep & 0.9299 & 0.9013 & 0.9233 & 0.8723 & 0.8951 & 0.8423 & 0.8302 & 0.8849  \\
   \hline\hline
        ours & \textbf{0.9438} & 0.9021 & \textbf{0.9323} & 0.8628 & \textbf{0.9357} & \textbf{0.8806} & \textbf{0.9400}  & \textbf{0.9139} \\
    \bottomrule
    \end{tabular}%
  \label{tab:addlabel}%
   }
\end{table}%

\noindent\textbf{The domain adaptation capability of  \abbr{}.} 
We verify the domain adaptation capability on  Optic Disc/Cup segmentation tasks. In an unseen client, we extract and freeze their encoders trained on other clients, only fine-tuning the decoders. In Table 7,  \abbr{} obtains the optimal result. (More relevant experiments in supplementary materials.)

\section{Conclusion}
Our proposed  \abbr{} addresses features challenge faced in personalized cross-silo federated medical image segmentation. \abbr{} incorporates Local Feature Enhancement and utilizes decoder as the personalized layer, ensuring both convergence and performance stability. We perform two-stage training and aggregation on global part (encoder), effectively improving its generalization and robustness. This enables it to adapt to new domains on unseen clients. Extensive experiments demonstrate superiority of our approach over existing personalized methods, approaching the performance of centralized learning. 

\subsubsection{Acknowledgements.} This work was supported by the National Key R$\&$D Program of China under Grant No.2022YFB2703301.

\subsubsection{Disclosure of Interests.} The authors have no competing interests to declare that are relevant to the content of this article.

\end{document}